\documentclass{article}
\usepackage{graphicx} 
\usepackage{subcaption}
\usepackage{listings}
\usepackage{xcolor}
\usepackage{authblk}
\usepackage{bookmark}
\usepackage{rotating}
\usepackage{lscape}
\usepackage{float}
\PassOptionsToPackage{table}{xcolor}
\usepackage[table]{xcolor}
\usepackage{longtable}
\usepackage[title]{appendix}
\usepackage[font=small,labelfont=bf]{caption}
\usepackage{adjustbox}
\usepackage{booktabs}
\usepackage{changepage}
\usepackage{array}
\usepackage{multirow}
\usepackage{MnSymbol}
\usepackage{placeins}
\usepackage{amsmath}

\usepackage[margin=1.4in]{geometry}

\usepackage[numbers]{natbib}
\title{A Graph Neural Network deep-dive into successful counterattacks}
\author[1,2,3]{Joris Bekkers}
\author[1]{Amod Sahasrabudhe}

\affil[1]{U.S. Soccer Federation}
\affil[2]{PySport}
\affil[3]{UnravelSports}

\date{October 2024}

\begin{document}

\maketitle

\begin{abstract}
A counterattack in soccer is a high speed, high intensity direct attack that can occur when a team transitions from a defensive state to an attacking state after regaining possession of the ball. The aim is to create a goal-scoring opportunity by covering a lot of ground with minimal passes before the opposing team can recover their defensive shape. The purpose of this research is to build gender-specific, first-of-their-kind Graph Neural Networks to model the likelihood of a counterattack being successful and uncover what factors make them successful in both men's and women's professional soccer. These models are trained on a total of 20,863 frames of algorithmically identified counterattacking sequences from synchronized StatsPerform on-ball event data and SkillCorner spatiotemporal (broadcast) tracking data. This dataset is derived from 632 games of MLS (2022), NWSL (2022) and international women’s soccer (2020-2022). With this data we demonstrate that gender-specific Graph Neural Networks outperform architecturally identical gender-ambiguous models in predicting the successful outcome of counterattacks. We show, using Permutation Feature Importance, that byline to byline speed, angle to the goal, angle to the ball and sideline to sideline speed are the node features with the highest impact on model performance. 

Additionally, we offer some illustrative examples on how to navigate the infinite solution search space to aid in identifying improvements for player decision making.  

This research is accompanied by an open-source repository containing all data and Python code hosted on the official U.S. Soccer Federation GitHub. It is also accompanied by an open-source Python package, $unravelsports$, which simplifies converting spatiotemporal data into graphs, and facilitates testing, validation, training, and prediction with this data. This should allow the reader to replicate and improve upon our research more easily.

\end{abstract}

\section{Introduction}
\label{intro}
Over the past couple of years, the engagement with the women’s game has increased significantly compared to years prior. The United States 2-0 victory over the Netherlands in the 2019 World Cup final became the most watched Women’s World Cup match ever with an increased viewership of 56\% compared to the 2015 final \cite{fifa2019}. Live viewers for the 2022 EURO increased by more than 50\% compared to 2017 and 214\% compared to 2013 \cite{uefa}. The NWSL reported a 71\% increase in viewers for the 2022 Championship final compared to 2021 \cite{nwsl}. Not only did the women’s game make these great strides forward on engagement, but it also did financially. FIFA more than tripled the prize money for the 2019 World Cup to \$50 million (from \$15 million in 2015) \cite{fifa20192}. In 2022 U.S. Soccer, the USWNTPA and the USNSTPA agreed to a collective bargaining agreement that achieved equal pay through identical economic terms for both senior national teams \cite{ussf}. And the creation of more women’s national competitions has ensured more women can play professionally.  

This increase in fan engagement and financial opportunity should be followed by an increase in data availability and subsequent data analysis to add additional support for the growth of the women’s game. Unfortunately, existing research into women’s soccer tactical decision-making is still scarce, but researchers have successfully mapped differences between genders in shooting tendencies \cite{bransen21, pappalardo, worville} and \cite{statsbomb2} used women’s event and freeze-frame positional data to profile passing in different competitions. To the best of our knowledge, no research exists to date that leverages tracking data to analyze differences in tactical tendencies between the men’s and women’s game. 

\subsection{Research}
\label{research}
This research is the first of its kind to use a full season of women’s and men’s professional soccer spatiotemporal broadcast tracking data (synced with on-ball event data). We use this data to train gender-specific state-of-the-art binary classification Graph Neural Networks (GNNs) with the aim of predicting the successful outcome of counterattacks.  

Counterattacks (or attacking transitions) are an important part of soccer strategy, and they play a prominent role in the U.S. Soccer Federation’s internal style of play guide.  

In general, a counterattack in soccer can be defined as a high speed, high intensity direct attack that can occur when a team transitions from a defensive state to an attacking state after regaining possession of the ball. The aim is to create a goal-scoring opportunity by covering a lot of ground with minimal passes before the opposing team can recover their defensive shape \cite{osmanbasic}.  

As per our algorithm described in Section~\ref{method}, 7.5\% of all shots attempted came from a counterattack in the 2022 MLS season. These counterattacking shots lead to 9.7\% of goals scored. Similarly, in the 2022 NWSL season 6.2\% of total shots came from counterattacks. These shots lead to 9.5\% of all goals scored during that season. 

\subsection{Graphs}
\label{graphs}
Graph Neural Networks rely on graph representations of frames of soccer data to capture the relevant information within it. In graph theory, a graph consists of a set of points (nodes) linked together by lines (edges). Historically, on-ball event data has been used to analyze \textit{The Beautiful Game} with such graphs by describing team, match and/or season aggregate information to represent players as the nodes of the graph and passes as the edges between these nodes, because the on-ball event data does not contain information of players off the ball. Researchers have used these graph representations of soccer to describe games in the form of pass networks \cite{bekkers17, mullenberg}, create adjacency matrices of all passes in a team to calculate the importance of players in a pass network \cite{clemente15} and flow motifs to identify unique playing styles from passing behavior of teams and players \cite{bekkers19, gyarmati14, pena12}.  

\begin{figure}[!ht]
    \centering
    \includegraphics[width=0.4\linewidth]{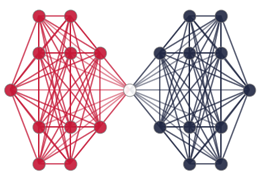}
    \caption{Schematic stylized graph representation of a single frame of tracking data}
    \label{fig:fig1}
\end{figure}

With the advent of (broadcast) tracking data – consisting of x, y coordinates of all players and the ball at 10 or 25 frames per second – it becomes possible to represent individual frames of tracking data with a similar graph representation. In these graphs, the players are still the nodes, but the edges now simply describe the relationship between all players on the pitch during a particular frame. Figure~\ref{fig:fig1} depicts a schematic, stylized representation of two teams (red and blue) connected to all their teammates and to the opposing team through the ball (in the center). 

Most contemporary research that leverages neural networks in soccer use a range of different image representations of the player and ball coordinates \cite{bauer23, fernandez21, statsperfrom} to communicate the information in a frame of soccer data to the neural network, because conventional convolutional neural networks are built to process images, not graphs. Starting only a few years ago, when GNNs started to become more mainstream due to their integration within machine learning packages \cite{fey19, grattarola21, hu2020open}, did we see researchers try their hand at representing on-ball event data \cite{minogue} and coordinate tracking data \cite{friends, stockl21, xenopoulos21} as graphs. 

Representing frames of tracking data as graphs allows us to not only incorporate the position of the players, but it makes it considerably simpler to add information to each node of the graph about a player’s speed, acceleration, distance to goal or even their preferred shoe color if we thought that mattered. Similarly, the edge features of the graph contain information on the connections between players, such as the inter-player distances or the inter-player angles. 

The ability to add this level of granular detail to the models allows us to manage the issues raised in \cite{statsperfrom} such as incomplete frames and games with less than 22 players on the pitch (due to red cards, injuries, or data inconsistencies etc.). Because we do not have to convert the set of coordinates to an image (and thereby losing valuable and interpretable information), and because we can include more detail into each graph, we can get a better understanding of which features have an impact on the model performance.  

\section{Methodology}
\label{method}
In this research we build different GNNs on three datasets (a women’s dataset, a men’s dataset, and a combined dataset) with the aim of predicting the outcome (successful or unsuccessful) of sequences of play considered a counterattack according to a rules-based algorithm. Subsequently, we use the different models built with these datasets to learn more about the differences between the men’s and women’s game by calculating the impact each of the individual features has on the model accuracy. Finally, we dive more deeply into some example situations to see what situationally dependent adjustments players could have made to improve their chances of successfully completing a counterattack according to our models. 

To accomplish this, we use synchronized StatsPerform on-ball event data, and SkillCorner broadcast spatiotemporal (10Hz) tracking data of a total of 632 recent games from the National Women’s Soccer League, International Women’s Friendlies, SheBelieves Cup, Olympic Women’s Tournament and Major League Soccer.  

Due to the nature of broadcast tracking data some players will periodically be out of view of the camera. Even though GNNs can deal with different amounts of players in different graphs we use SkillCorner’s predicted coordinates for these out-of-view players to be able to have 22 players in most frames.  We ensure the models are built on good quality data by only using games that have a SkillCorner player and ball quality ratings of 4 out of 5 or above. 

To build a Graph Neural Network model to predict the successful outcome of counterattacks requires us to first identify phases of play that can be labeled as part of a counterattack in both the men’s and women’s game. Because \cite{bauer21} has shown us how manually labeling specific situations in soccer can be a time intensive task, we have opted to algorithmically identify counterattacks and subsequently label them successful or unsuccessful using a set of on-ball event driven rules comprised and verified in collaboration with USWNT and USMNT Performance Analysts, and in accordance with the U.S. Soccer Federation’s internal style of play guide. 

Though the exact set of rules can be found in the Appendix, it is important to note – due to the limited number of goal attempts in our dataset – we consider a successful counterattack to end with the attacking team moving the ball in the opposing team’s penalty area, either via a successful on-ball run, or by receiving the ball in the box successfully. 

The algorithmic approach yields a set of sequences of on-ball events considered to be a counterattack. We use this set of sequences to label all individual frames of SkillCorner broadcast tracking data as part of a successful counterattack (1) or part of an unsuccessful counterattack (0) and we omit all frames not considered part of a counterattack.  

This means that our model only uses individual frames (without timeseries components), and each frame is labeled with the future outcome (successful or unsuccessful) of the sequence within which it lies.  

In the on-ball event data model VAEP (Valuing Actions by Estimating Probabilities) \cite{decroos20} a similar forward-looking labeling technique is used, but instead each on-ball action is annotated when a goal is scored or conceded within 10 actions from the current action.  

\subsection{Model Architecture}
\label{architecture}
The individual raw tracking data frames are converted into a graph representation to train the GNNs with the Python library \textit{Spektral} \cite{grattarola21}. The graph representation consists of individual matrices with node features, edge features and an adjacency matrix for each frame. The node features are comprised of normalized player’s coordinates, velocity, angle of motion, distance to goal, angle to goal, distance to the ball, angle to the ball and an attacking team flag. The edge features consist of inter-player angles (in the form of a sine and cosine value) and normalized inter-player distances. Within the adjacency matrix players from the same team are connected to each other and every player is connected to the ball node, as shown in Figure~\ref{fig:fig1}.  

\begin{figure}[!ht]
    \centering
    \includegraphics[width=0.75\linewidth]{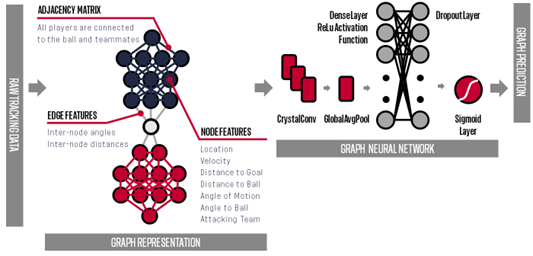}
    \caption{Graph Neural Network model architecture}
    \label{fig:fig2}
\end{figure}

Figure~\ref{fig:fig2} depicts the GNN model architecture. It contains three CrystalConv layers \cite{xie18} (a method originally built to predict crystalline properties using connections of atoms) to directly learn game state properties from the connections between players in each graph, a Global Average Pool layer, a Dense layer with ReLu activation, a Dropout layer and ultimately a Sigmoid activation function.

\subsection{Dataset}
\label{dataset}
The GNNs are trained on a balanced training set (consisting of 70\% of samples) containing 50\% successful, and 50\% unsuccessful counterattacks. One model is trained on only women’s data, one on only men’s data and one combined model trained using all men and women’s data. Due to the lower number of samples available for the women’s data (due to fewer games), the models trained using the women’s data are trained on 100 epochs whereas the other models are trained on 200 epochs. Table~\ref{data} describes the data used for training these models.  

\begin{table}[!ht]
    \centering
    \begin{tabular}{lccc}
        \toprule
        & \textbf{Women} & \textbf{Men} & \textbf{Combined} \\
        \midrule
        \textbf{Competitions} & 
        \begin{tabular}[t]{@{}c@{}}NWSL \\ Int. Friendlies \\ SheBelieves Cup \\ Olympics \end{tabular} & 
        \begin{tabular}[t]{@{}c@{}}MLS\end{tabular} & 
        \begin{tabular}[t]{@{}c@{}}NWSL \\ Int. Friendlies (W) \\ SheBelieves Cup \\ Olympics (W) \\ MLS\end{tabular} \\
        \midrule
        \textbf{Games} & 157 & 475 & 632 \\
        \textbf{Counterattacks} & 942 & 3782 & 4727 \\
        \textbf{Counterattack frames} & 3720 & 17143 & 20863 \\
        \bottomrule
    \end{tabular}
    \caption{Data breakdown}
    \label{data}
\end{table}

\subsection{Open Source}
\label{open}
In the interest of expanding the research conducted using women’s spatiotemporal data, to give others an opportunity to build on top of our models and more importantly to try to improve them, we have made both the dataset (in its anonymized graph representation form as used in this research), and an interactive Python Jupyter Notebook used to train our models available for download on the U.S. Soccer Federation’s Official GitHub page \cite{ussfgh}.  
Additionally, the open-source Python package $unravelsports$ \cite{unravelsports} has been released to aid researchers in converting spatiotemporal data into Graphs and subsequently training, validating, testing and predicting using built-in support for the $Spektral$ package \cite{grattarola21}.

We encourage readers to try to improve our models’ results by including different node and edge features and by trying out different types of adjacency matrices (all of which have been included in the downloadable datasets) in addition to examining different model architectures. 

\section{Results}
\label{results}
To evaluate our model performance, we compare the gender-specific models to a naive baseline model that assumes a 50/50 class split (resulting in a naive Log-Loss of 0.693 and an ROC-AUC of 0.50) and we compare it to the model trained on the combined dataset.  

Table~\ref{performance_metrics} shows the Log-Loss and ROC-AUC scores for the naive model, the combined model, and the gender-specific models. The lower Log-Loss and the higher ROC-AUC scores for the gender-specific models indicate that it pays to train gender-specific models to improve model performance, and it can achieve this with a significantly smaller sample size (as shown in Table~\ref{data}). 

The seemingly high Log-Loss for all models can, for the most part, be attributed to the predictions made on frames where play is concentrated in the center of the pitch. In this area, predictions mostly lie within the 40-60\% success range, meaning that we are essentially trying to predict the outcome of a coin-flip. With this in mind, we do believe that even though our model architecture can most likely be improved, our models already performed well given the difficulty of the prediction problem. 

\begin{table}[!ht]
    \centering
    \begin{tabular}{lcccc}
        \toprule
        & \textbf{Women} & \textbf{Men} & \textbf{Combined} & \textbf{Naive} \\
        \midrule
        \textbf{Testing Log Loss} & 0.48 & 0.51 & 0.56 & 0.69 \\
        \textbf{ROC-AUC} & 0.83 & 0.78 & 0.76 & 0.50 \\
        \bottomrule
    \end{tabular}
    \caption{Model performance metrics}
    \label{performance_metrics}
\end{table}

\subsection{Model Calibration}
\label{calib}
Due to the pseudo-probabilistic nature of the predictions, they can only aid in our understanding of the game if they are well-calibrated. We can measure the calibration by calculating the Expected Calibration Error (ECE). The ECE gives a weighted average over the difference between the absolute accuracy and the confidence, calculated per probability bin. The ECE values for the men’s and women’s model are 0.15 and 0.18 respectively. 

\begin{figure}[!ht]
    \centering
    \includegraphics[width=0.4\linewidth]{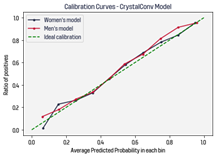}
    \caption{CrystalConv model calibration curves}
    \label{fig:fig3}
\end{figure}

The Calibration Curves (displayed in Figure~\ref{fig:fig3}) represent the distribution of the predicted probabilities for the men’s model and women’s model.  

The ECE values and the Calibrations curves clearly indicated that our models are indeed well calibrated.  

\subsection{Feature Importance}
\label{fi}
The increased model performance in both the men and women’s model compared to the combined model indicates that these gender-specific models must weigh their respective model’s features differently to achieve these improved results. We use Permutation Feature Importance \cite{altmann} to calculate the feature importance within both gender’s models to uncover where these differences lie. Permutation Feature Importance allows us to identify the importance of each individual feature by measuring the increase in prediction error when breaking the relationship between individual features and the observed result through the application of a random permutation to the feature’s values. In other words, we use the test set to randomly shuffle the values for one feature and recalculate the model error \cite{molnar}. Next, we return those shuffled values to their true state and repeat the process for every other feature individually.  

\begin{figure}[!ht]
    \centering
    \includegraphics[width=0.9\linewidth]{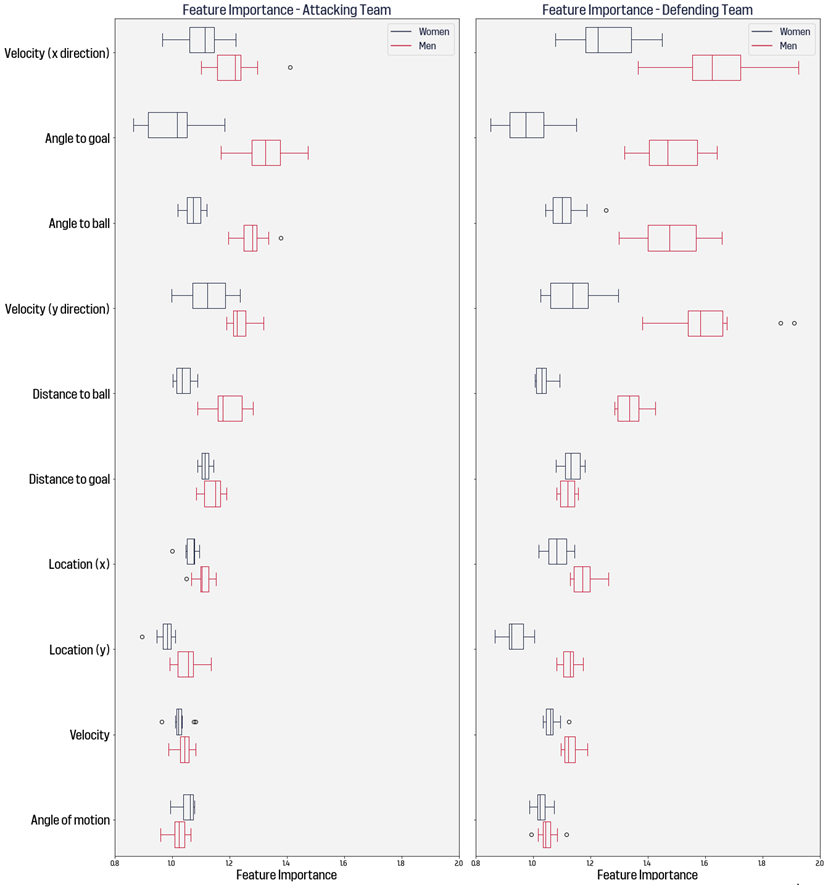}
    \caption{Attacking and Defending Feature Importance for the Women's and Men's models}
    \label{fig:fig4}
\end{figure}

Applying multiple independent random shuffles and calculating the model’s error for each individual shuffle allows us to analyze both the average and the spread of the acquired model errors for each feature. 

The feature importance is measured by subtracting the AUC obtained after randomly shuffling the values for a feature from the AUC of the actual model.  

In Figure~\ref{fig:fig4} we show the results of applying this Permutation Feature Importance random shuffle to each of the 10 node features 15 times, both for the attacking players (on the left) and for the defending players (on the right) for both gender-specific models. Because it is impossible to shuffle the edge features without completely breaking the logic of the individual graphs, we have left those out of the scope of this feature importance analysis. 

In Figure~\ref{fig:fig4} we see that the x-component of the velocity vector (in other words, the players’ speed from byline to byline) and the players’ angle to the goal are the two features with the highest impact on our model’s performance. The ball angle between each player and the ball, the y-component of the velocity vector (sideline to sideline speed) and each player’s distance to the ball are also relatively important. Finally, we see how the x and y coordinates, the speed and the direction of movement are the least impactful. This can be partially attributed to the fact that these values are highly correlated with the distance and angle to the goal, and the components of the velocity vector respectively. Finally, notice how the defending team’s features are more important to the model’s performance than the attacking players features. This could be because, by definition, a counterattack exists solely during a phase of disorganized defending. 

\section{Application}
\label{application}
\begin{figure}[!ht]
    \centering
    \begin{subfigure}[b]{0.4\textwidth}
        \centering
        \includegraphics[width=\linewidth]{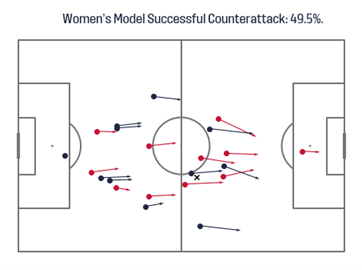}
        \caption{\centering The women's model predicts this situation to end in the opponents box 49.5\% of the time.}
        \label{fig:fig5}
    \end{subfigure}
    \hfill
    \begin{subfigure}[b]{0.4\textwidth}
        \centering
        \includegraphics[width=\linewidth]{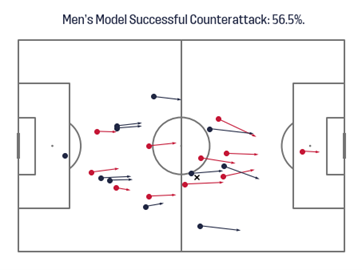}
        \caption{\centering The men's model predicts this situation to end in the opponents box 56.5\% of the time.}
        \label{fig:fig6}
    \end{subfigure}
    \caption{\centering An example prediction on the same situation using both models. \newline \footnotesize The blue team is attacking left to right}
    \label{fig:side_by_side1}
\end{figure}

The impact of the differences between these gender-specific models on an aggregated node feature level is challenging to translate into actionable insights from a figure like Figure~\ref{fig:fig4}, nor are they easily understood (at this time) by visually inspecting frames and comparing the differences to their respective counterattack success probability like in Figure~\ref{fig:side_by_side1}. More research is needed to translate the information from Figure~\ref{fig:fig4} into actionable insights, and to connect that knowledge into a direct understanding of the probabilistic differences between Figure~\ref{fig:side_by_side1}.  

The importance of these gender-specific models at the time of writing lies in the opportunity to offer a Women’s team’s coaching staff the ability to interact with a model that is trained specifically on women’s data, because currently most, if not all, models are trained primarily on men’s data. It is therefore better suited to offer input for analysis and visual inspection of sequences of play relevant to the women’s game. 

\subsection{An Infinite Search Space}
\label{infinite}
The models created for this research can be highly valuable when they are built into an interactive tool such as \cite{bekkers22}. Tools like this allow coaches, technical staff, analytics staff, and players alike to interactively navigate the infinite optimization search space of individual frames or even specific sequences of play. This can help them to find small improvements or novel solutions for their tactical approaches and their positioning and movement decisions, both from an attacking and a defending perspective.  

\begin{figure}[!ht]
    \centering
    \begin{subfigure}[b]{0.4\textwidth}
        \centering
        \includegraphics[width=\linewidth]{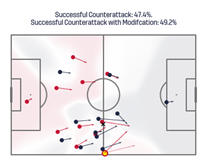}
        \caption{An improved run trajectory for the right winger up to 49.2\% from 47.4\%}
        \label{fig:fig7}
    \end{subfigure}
    \hfill
    \begin{subfigure}[b]{0.4\textwidth}
        \centering
        \includegraphics[width=\linewidth]{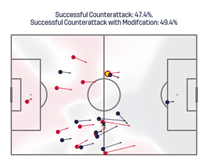}
        \caption{An improved run trajectory for the left winger up to 49.4\% from 47.4\%}
        \label{fig:fig8}
    \end{subfigure}
    \caption{\centering Improved run trajectories for both wingers. \footnotesize The faint arrow represents the original trajectory of the player. Pitch control added for visual guidance only}
    \label{fig:side_by_side2}
\end{figure}

\begin{figure}[!ht]
    \centering
    \includegraphics[width=0.4\linewidth]{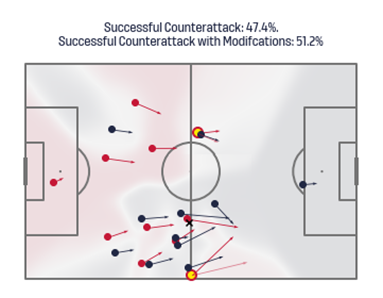}
    \caption{\centering An improved run trajectory for both the right winger and left winger. Successful counterattack probability up to 51.2\% from 47.4\%. \footnotesize The faint arrow represents the original trajectory of the player. Pitch control added for visual guidance only}
    \label{fig:fig9}
\end{figure}

\begin{figure}[!ht]
    \centering
    \begin{subfigure}[b]{0.4\textwidth}
        \centering
        \includegraphics[width=\linewidth]{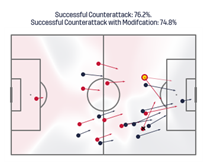}
        \caption{\centering A worse run trajectory down to 74.8\% from 76.2\%}
        \label{fig:fig10}
    \end{subfigure}
    \hfill
    \begin{subfigure}[b]{0.4\textwidth}
        \centering
        \includegraphics[width=\linewidth]{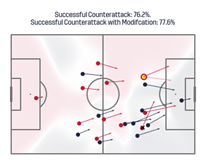}
        \caption{\centering An improved run trajectory up to 77.6\% from 76.2\%}
        \label{fig:fig11}
    \end{subfigure}
    \caption{\centering A comparison of run trajectories for the left winger. The faint arrow represents the original trajectory of the player. Pitch control added for visual guidance only}
    \label{fig:side_by_side3}
\end{figure}

Figure~\ref{fig:side_by_side2} shows an example of a search for an improved attacking run during a counterattack. To offer some additional visual guidance we have reinforced the figure with a Pitch Control model \cite{spearman17}. To find this improvement (and the improvements shown in Figure~\ref{fig:fig8} ~\ref{fig:fig9}, ~\ref{fig:fig10} and ~\ref{fig:fig11}) we have changed the highlighted player’s movement direction by increments of 15° to uncover a potentially improved run trajectory. Finding these improvements in practice could be accomplished with a computer-assisted interactive search tool. 

When the right winger’s trajectory in Figure~\ref{fig:fig7} is rotated by 30° to the inside the probability of successfully completing this counterattack increases by 1.8 percentage points.  

Using the same frame again in Figure~\ref{fig:fig8} but now rotating the left winger’s trajectory outwards by 30° increases the prediction by 2 percentage points. In this case, running away from the direction of the defensive player, and running into a channel with more space, increases the likelihood of a successful counterattack. Figure~\ref{fig:fig9} shows how both improvements simultaneously can help increase the success probability by 3.8 percentage points. 

In the next on-ball action, the right winger has taken control of the ball. The left winger can now choose to position more centrally as shown in Figure~\ref{fig:fig10}. This results in the model prediction dropping down by 1.4 percentage points, because that run is directed towards a part of the pitch which would be dominated by the opposition team. 

On the other hand, a run in the wide channel as depicted in Figure~\ref{fig:fig11}, will lead to the highlighted player occupying more open space. This leads to an increase of 1.4 percentage points in the probability of successfully completing the counterattack. 

In addition to searching for these run optimizations, these models can be used to evaluate both player and team performance in counterattacks, to uncover players that help reduce their team’s likelihood of conceding a successful counterattack, or player that can improve their team’s chance of completing a successful counterattack via their positioning or their run selection.

\section{Discussion}
\label{discussion}
Within this research we have built and open-sourced two gender-specific Graph Neural Network models to predict the outcome of counterattacks, leveraging individual frames of spatiotemporal tracking data by algorithmically assigning labels to them from on-ball event data. 

In comparison to Expected Possession Value (EPV) models \cite{fernandez22} build using conventional convolutional neural networks, our GNN models can incorporate more features, such as speed and direction of motion. A major difference between this EPV model and our models is the use of the penalty area as the target for successfully completing a counterattack. We would have preferred to use (expected) goals (or goal attempts) as the successful end-product; however, we did not have enough samples to satisfy this preference. This means that, for example, the improvements found in Section~\ref{infinite} will help us control the ball inside the box, but they might also reduce the chance of scoring a goal. 

In contrast to on-ball event data-based models, our implementation allows us to value each individual player’s contribution, both on and off the ball, primarily due to the nature of the spatiotemporal data. But, contrary to VAEP it does not evaluate the risk of conceding a new counterattack for the current attacking team. Our output is simply an assessment of the threat going forward, like Expected Threat \cite{singh}. 

Additionally, we need to acknowledge that even though we attempted to reduce the risk of over-fitting (especially on the smaller women’s dataset) by reducing the number of epochs during training this research certainly lacks a thorough validation of the model performance. 

Furthermore, we need to accept that the use of an on-ball event-based algorithm to identify counterattacks is not a perfect solution. However, we feel that manually annotating counterattacks for both genders is perhaps even more prone to introducing biases into the training sample. 

To extend this work we would like to conduct more research on the interpretation of the feature importances, and how they can be narrowed down further into actionable insights. Supplementing our approach with a model that predicts the likelihood of a counterattack starting given all the players’ positioning, speed and directionality values could help us streamline the model’s usage.  

Unfortunately, after finalizing this research we became aware of \cite{statsbomb}. This Expected Goals research conducted by StatsBomb describes a gender-aware approach.  In this third option – gender-specific and a simple combined setup being the other two – a feature is added to their tabular data to indicate the gender of the player taking the shot. Even though this seems obvious in hindsight, due to the non-tabular nature of our GNN datasets, it was not immediately evident to us to attempt to add such a variable at the time of conducting our research. As a result, we suggest future gendered GNN research incorporated a female/male variable (as a node features) to uncover if this makes a difference. 

Finally, to facilitate more public research into gendered counterattacking GNNs we have released an additional (imbalanced) Graph dataset on GitHub \cite{ussfgh}. This dataset consists of 210,000 frames belonging to counterattacking phases of play split equally between women and men. Approximately 5\% of frames in this dataset are labeled a success as they belong to counterattacks that lead to a goal. 

\section{Conclusion}
\label{conclusion}
We have demonstrated that it is possible to build gender-specific Graph Neural Network models that outperform gender-ambiguous models in predicting the successful outcome of counterattacks while relying on a significantly smaller sample size (especially for the Women’s model). With the help of Permutation Feature Importance, we have shown that byline to byline speed and angle to the goal have the biggest impact on model performance for both genders, in attack and defense. Using the same technique, we have also uncovered that the defending players’ node features have more impact on model performance than of their attacking counterparts. To better understand these implications, we propose an interactive tool to help navigate the infinite movement, speed, and positioning search space to aid players and coaching staff in finding small improvements or novel solutions for their tactical approaches and their positioning and movement decisions, both from an attacking and a defending perspective. Finally, we challenge the reader to improve our model architecture and the choices we have made for the node and edge features, by using the models and data provided in the U.S. Soccer Federation GitHub repository, and to use the $unravelsports$ package to more easily apply the techniques discussed in this paper to bespoke data.

\begin{appendices}
\end{appendices}

\end{document}